\documentclass{article}

\PassOptionsToPackage{numbers}{natbib}
\usepackage[final]{neurips_2024}

\usepackage[utf8]{inputenc} % allow utf-8 input
\usepackage[T1]{fontenc}    % use 8-bit T1 fonts
\usepackage{hyperref}       % hyperlinks
\usepackage{url}            % simple URL typesetting
\usepackage{booktabs}       % professional-quality tables
\usepackage{amsfonts}       % blackboard math symbols
\usepackage{nicefrac}       % compact symbols for 1/2, etc.
\usepackage{microtype}      % microtypography
\usepackage{xcolor}         % colors
\usepackage{algpseudocode}
\usepackage{algorithm}
\usepackage{amsmath}
\usepackage{amsthm}
\usepackage{graphicx}
\usepackage[title]{appendix}

\newtheorem{theorem}{Theorem}

\title{Causal Order Discovery based on Monotonic SCMs}

\author{%
  Ali Izadi \\
  Department of Computing Science\\
  Simon Fraser University\\ 
  \texttt{ali\_izadi@sfu.ca} \\
  \And
  Martin Ester \\
  Department of Computing Science\\
  Simon Fraser University \\
  \texttt{ester@sfu.ca} \\
}

\begin{document}

\maketitle

\begin{abstract}

In this paper, we consider the problem of causal order discovery within the framework of monotonic Structural Causal Models (SCMs), which have gained attention for their potential to enable causal inference and causal discovery from observational data. While existing approaches either assume prior knowledge about the causal order or use complex optimization techniques to impose sparsity in the Jacobian of Triangular Monotonic Increasing maps, our work introduces a novel sequential procedure that directly identifies the causal order by iteratively detecting the root variable. This method eliminates the need for sparsity assumptions and the associated optimization challenges, enabling the identification of a unique SCM without the need for multiple independence tests to break the Markov equivalence class. We demonstrate the effectiveness of our approach in sequentially finding the root variable, comparing it to methods that maximize Jacobian sparsity. 
\end{abstract}

\section{Introduction}
The problem of finding the causal relationships among a set of variables, or causal discovery, has been extensively studied to address interventional and counterfactual queries (causal inference) \citep{Scholkopf2017-da}. These two type of queries play a crucial role in biology \citep{zanga2022survey, imbens2004nonparametric}, economics \citep{imbens2004nonparametric}, philosophy \citep{malinsky2018causal}, and machine learning \citep{kaddour2022causal}. Structural Causal Models (SCMs) are among the most well-known frameworks for mathematically formulating the problem of causal discovery \citep{pearl2009causality}. SCMs model causal relationships by assigning each effect variable as a function of its causes and an unobserved noise. They enable the answering of interventional and counterfactual queries, as well as the learning of causal relationships from observational data. 

In general, one needs interventional data to learn the true, unique SCMs. However, for special types of SCMs the task of causal discovery is possible from just observational data. The identifiability of SCMs has been investigated by proposing different functional assumptions between causes and effect. These types of SCMs, including those with linear functions with non-Gaussian noise \citep{shimizu2006linear}, nonlinear functions with additive independent noise \citep{hoyer2008nonlinear}, and post-nonlinear models have been proven to be identifiable from observational data \citep{10.5555/1795114.1795190}. 

Recent research has demonstrated that Monotonic SCMs can be identified using triangular monotonic increasing (TMI) maps, provided the causal order of these maps is known \citep{xi2023indeterminacy}. Monotonic SCMs are a class of SCMs characterized by nonlinear functions between causes and effects, where each function is monotone with respect to unobserved noise. Given the true order of the maps, the Jacobian of normalizing flows transformation as TMI maps \citep{rezende2015variational} has been shown to effectively learn Monotonic SCMs \citep{javaloy2024causal}. Although knowing the order is a restrictive assumption for causal discovery, normalizing flows have been employed for causal inference \citep{javaloy2024causal}. Nevertheless, TMI maps can facilitate causal discovery based on Monotonic SCMs without requiring the order \citep{xi2023triangular}. It has been established that the true order of TMI maps corresponds to an order where the Jacobian of TMI maps is maximally sparse \citep{xi2023triangular}. This insight leads to a two-step optimization procedure for estimating TMI maps with sparse Jacobians at each iteration of searching over orders (permutations). This introduces significant complexity to the optimization efficiency of this procedure, with many local minima, as discussed in \citep{xi2023triangular}, without proposing an efficient method to overcome this issue. However, this approach identifies only the Markov Equivalent Class (MEC) of the true SCM, and multiple independence tests need to be performed to identify the unique SCM.

In this paper, we propose a procedure for learning the causal order of Monotonic SCMs based on the idea of sequentially identifying the root of the graph. For Monotonic SCMs, the Jacobian vector of the map of the root variable with respect to all other variables contains all zero values. Instead of enforcing sparsity on the Jacobian of the entire TMI maps, we determine the order by sequentially discovering the roots using the zeroness of the Jacobian of the root map. This approach enables the learning of the true order of the SCMs, making SCMs identifiable \citep{xi2023indeterminacy} and breaking the Markov Equivalent Class (MEC), eliminating the need for independence tests. Additionally, this sequential procedure, without any sparsity assumption, simplifies the complex two-step process of learning TMI maps and orders. In summary, the main contributions of this paper are as follows:

\begin{itemize}
     \item We introduce a novel sequential procedure for discovering causal order in Monotonic SCMs by leveraging the Jacobian of TMI maps to identify the root variable, and to the best of our knowledge, this represents the first efficient approach for causal discovery based on Monotonic SCMs.
     
     \item We conduct several experiments to demonstrate the efficiency of our approach in finding the correct order by sequentially discovering the root variable. Our method is compared to approaches that aim to learn the order with the maximally sparse Jacobian.
     
\end{itemize}

\section{Related Works}
\textbf{Nonlinear SCMs}:
The approach to causal discovery for nonlinear SCMs using the Jacobian began with \citep{zheng2020learning}, but it has focused on additive nonlinear noise models rather than monotonic SCMs. In contrast, the connection between nonlinear ICA and causal discovery for nonlinear SCMs has been explored in \citep{reizinger2023jacobian, monti2020causal}. However, identifying the true unique graph through ICA requires counterfactual data, which is a restrictive assumption compared to our proposed method that relies on observational data. While the identifiability of monotonic SCMs was introduced through the concept of indeterminacy in generative models \citep{xi2023indeterminacy}, knowing the causal order is essential. Our paper, however, focuses on determining this causal order for monotonic SCMs. Additionally, the assumption of bijectivity for nonlinear SCMs has been demonstrated to be useful for counterfactual identifiability results in Monotonic SCMs.
A recent fixed-point approach for causal discovery based on nonlinear SCMs involves amortized causal order discovery, but does not specifically address monotonic SCMs \citep{scetbonfixed}.

\textbf{Causal order discovery}:
Many approaches in causal discovery focus on searching the space of causal orders rather than the space of DAGs. These methods first identify the topological order of the causal graph and then apply an edge pruning algorithm to obtain the final causal graph \citep{peters2014causal, buhlmann2014cam}. RL-based \citep{wang2021ordering} and Permutation based \citep{raskutti2018learning, squires2020efficient} determine the causal order using a greedy approach, which is not applicable for high-dimensional data due to its combinatorial nature.

Another set of order-based methods attempts to find the topological order based on discovering permutation matrix \citep{charpentier2022differentiable, zantedeschi2023dag, cundy2021bcd}. While these methods make the problem of causal discovery fully differentiable by proposing a procedure for learning the combinatorial permutation matrix, their approaches suffer from searching in the space of permutation matrices, which is challenging to optimize due to its combinatorial nature. Another line of research focuses on using the score or gradient of probability distribution of data for causal order discovery \citep{rolland2022score, sanchez2022diffusion, montagna2023causal, montagna2023scalable, zhu2024sample}. In contrast to score-bases methods that leverage the variance of the Hessian of the probability distribution for selecting leaves of graph iteratively, our method uses Jacobian of TMI maps for finding the roots iteratively. However, unlike the proposed method, none of the methods assume Monotonic SCMs. 

% Finally, our work is similar to AbPNL \citep{uemura2022multivariate} and Rank-PNL \citep{keropyan2023rank} in terms of finding the causal order assuming post-nonlinear SEM. In contrast, our method employs the normalizing flow architecture and a masking mechanism in the training procedure to enhance computational efficiency for data with higher dimensions.  

\textbf{TMI maps and Normalizing flows for causal discovery}:
Flow-based models have been employed for causal discovery under the assumptions of location-scale noise models \citep{khemakhem2021causal, kamkari2023ocdaf}. Our method differs from these two, as we assume the monotonic SCMs. Another group of methods develops the theory of normalizing flows for causal inference and proposes an algorithm for causal discovery using normalizing flows \citep{javaloy2024causal}. However, this algorithm assumes that the topological order of the graph is given as input, whereas the focus of our paper is on discovering this causal order. 
TMI maps have also been used to learn conditional independence through sparsity, but this application is limited to undirected settings in graphical models, not for causal discovery \citep{spantini2018inference, morrison2017beyond, baptista2024learning}. While TMI maps have been applied to causal discovery, their use has been confined to identifying the MEC equivalence class based on conditional independencies derived from TMI maps \citep{akbari2023causal, xi2023triangular}. In contrast, our work focuses on discovering the causal order of the graph, without relying on conditional independencies or outputting the Markov equivalence class.

\section{Background}\label{sec:background}
\subsection{Monotonic Structural Causal Models}\label{MSCM}
Given a $d$-dimensional observational dataset $X \in \mathcal{R}^{n\times d}$, where X contains $n$ i.i.d. samples. The causal relationships among the set of variables are captured by a Directed Acyclic Graph (DAG) $\mathcal{G}$, which is represented by the adjacency matrix $A \in \{0, 1\}^{d \times d}$. The adjacency matrix $A$ models the causal relationship of each random variable $x_i$ and its parents $Pa(x_i)$. We can write the generative process of $x_i$ in the form of Equation \ref{eq:mscm}
\begin{equation}\label{eq:mscm}
    x_i = f_i(Pa_{\mathcal{G}}(x_i), u_i), \; \; \; \;  i=1, ..., d
\end{equation}
where $f_i$ is an arbitrary function, $Pa_{\mathcal{G}}(x_i)$ denotes the parents of variable $x_i$ in DAG $\mathcal{G}$, and the exogenous noise $u_i$ is independent of $Pa_{\mathcal{G}}(x_i)$. Also, the different noise variables $u_i$ are mutually independent, which can be arbitrary independent noise distributions. For Monotonic SCMs, we assume that each $f_i$ is monotone in $u_i$ \citep{xi2023indeterminacy, javaloy2024causal, xi2023triangular}.

\subsection{Normalizing flows and TMI maps}
Normalizing flows are a class of generative models that transform the probability distribution of a random variable $X \in \mathcal{R}^d$ into the probability distribution of a random variable $U \in \mathcal{R}^d$ through a series of invertible and differentiable transformations $T(x)$ \citep{papamakarios2021normalizing, irons2022triangular}. A key aspect of this transformation process is the use of Triangular Monotonic Increasing (TMI) maps. A TMI map is a function $\mathcal{R}^d \to \mathcal{R}^d$ that is both triangular—where each output component $T_i$ depends on the $x_{1:i}$ components of the input, as expressed by
\begin{equation}\label{ref:tmi}
T(x_1, x_2, \dots, x_d) = \left( T_1(x_1), T_2(x_1, x_2), \dots, T_d(x_1, x_2, \dots, x_d) \right),
\end{equation}
and monotonic increasing w.r.t $x_i$ \citep{bogachev2005triangular, carlier2010knothe}, ensuring that the transformation is invertible. The triangular structure of TMI maps allows for efficient computation of the inverse map and the Jacobian determinant, which are essential for calculating the likelihood in normalizing flows.

\section{Proposed Method}\label{sec:method}
\subsection{Problem definition}
The objective of this paper is to determine the causal order of variables (Refer to Appendix \ref{appendix:causal_order} for the formal definition of causal order). Although a DAG $\mathcal{G}$ may have multiple causal orderings, for any given causal ordering $\pi$, we can construct a fully connected DAG $\mathcal{G_{\text{full}}^{\pi}}$ where, for every pair of nodes $\pi(i)$ and $\pi(j)$, there is a directed edge $\pi(i) \rightarrow \pi(j)$ if and only if $i < j$  \citep{peters2014causal}. We define the set of true causal orderings $\Pi$ as:   
\begin{equation}
    \Pi = \{ \pi | \mathcal{G_{\text{full}}^{\pi}} \text{\; is a super-graph of \;} \mathcal{G} \}
\end{equation}
Under the assumption of a monotonic SCM (Section \ref{MSCM}), the set $\Pi$ is identifiable. Therefore, the goal of causal order discovery is to find a $\pi \in \Pi$ \citep{ng2022masked}.

\textbf{Assumptions}: i) Each function $f_i$ in Equation \ref{eq:mscm} is continuous, invertible, and therefore monotonic; ii) The graph is acyclic; iii) There are no hidden confounders, so the noise variables $u_i$ are mutually independent (causal sufficiency).

\subsection{Overview of the method}
Our proposed method finds the causal order of the nodes by iteratively identifying the roots of the graph, removing each root from the graph, and repeating this process until only one variable remains. The topological order of the graph is then determined as the order in which roots were found.
% Once the causal order is discovered, we employ the identifiability results of using causal normalizing flows \citep{javaloy2024causal} to extract the entire adjacency matrix $A$.

In the following sections, we detail our methodology for identifying the roots of the graph. This process involves training one-dimensional conditional normalizing flows to model the functions $f_i$ of a monotonic SCM, and using the Jacobian of each map ($\nabla T_i(x_i)$) as a criterion for detecting the graph's root. Next, we introduce an algorithm for sequentially finding the roots to determine the causal order. Once the causal order is discovered, one can either train a $d$-dimensional normalizing flow to identify the adjacency matrix \citep{javaloy2024causal}, or apply the CAM pruning method \citep{buhlmann2014cam} to prune the edges. This approach uniquely discovers the causal graph without relying on independence tests, unlike other methods that depend on the faithfulness assumption and the sparsity of the Jacobian to identify the Markov equivalence class and break it using independence tests \citep{xi2023triangular}.  

\subsection{Finding the root}

We assume $d$ independent multi-cause models, i.e. monotonic SCM, according to equation \ref{eq:multi-cause} as follows:
\begin{equation}\label{eq:multi-cause}
    x_i = f_i(X\backslash\{x_i\}, u_i), \; \; \; \;  i=1, ..., d
\end{equation}
where $X\backslash\{x_i\}$ are all variables except variable $x_i$, and $f_i$ are monotonic in $u_i$. Thus, $f_i^{-1}(X\backslash\{x_i\}, x_i)$ is also monotonic increasing w.r.t $x_i$, and it is a monotonic increasing map. As a result, for each $f_i^{-1}$ we train one-dimensional conditional normalizing flow $T_i(x_i, X\backslash\{x_i\})$, where each $T_i$ transforms $x_i$ to $u_i$ conditioned on $X\backslash\{x_i\}$ and each $T_i$ is a monotone increasing map for each $x_i$.  

\begin{theorem}\label{th:root}
Suppose $X \in \mathcal{R}^{n\times d}$ is generated based on a monotonic SCM according to Equation \ref{eq:mscm}. Then, for the multi-cause model \ref{eq:multi-cause}, $\nabla_{X\backslash\{x_i\}}T_i(x_i, X\backslash\{x_i\})=\mathbf{\vec{0}}$, if and only if $x_i$ is a root in $\mathcal{G}$.
\end{theorem}

\textbf{Sketch of the Proof}: The proof leverages Theorem 1 from \citep{javaloy2024causal}, which establishes that, under the assumption of a known causal order, the Jacobian of the transformation $T$ is $I - A$, where $A$ is a lower triangular matrix. Therefore, for the root variable $x_i$, the Jacobian $\nabla_{X\backslash\{x_i\}}T_i(x_i, X\backslash\{x_i\})=\mathbf{\vec{0}}$ is zero due to the structure of $A$. Refer to Appendix \ref{appendix:th} for a complete proof of Theorem \ref{th:root}.

To practically implement the criterion in Theorem \ref{th:root}, we detect the root by finding the maximum absolute value in each row of the Jacobian and selecting the variable corresponding to the row with the smallest of these maximum values.

\begin{equation}
    \text{Root variable} = \arg \min_{i=1,\dots,d} \max \left| \nabla_{X\backslash\{x_i\}} T_i(x_i, X\backslash\{x_i\}) \right|
\end{equation}

While other aggregation schemes, such as summation or Hotelling's T-squared test \citep{hotelling1931generalization}, can be applied here, we found the min-max criterion to be more stable in our experiments for detecting the root. 

\subsection{Causal order discovery}
Once the first root of the graph is detected, we remove the root variable from the current set of variables and add it as the first element in the causal order. We then continue finding subsequent roots until the entire topological order is determined.  Refer to Appendix \ref{appendix:algorithm} for a detailed description of the causal order discovery algorithm, as summarized in Algorithm \ref{alg:proposed}.

\section{Empirical evaluation}
\textbf{Synthetic data generation} \;
Following the previous works \citep{xi2023triangular, reizinger2023jacobian, monti2020causal}, the synthetic data was generated by first constructing a DAG with $d$ nodes, where edges were included with a probability of $\frac{2}{d-1}$. Edges were then randomly assigned to the nodes. Synthetic samples were drawn from a standard Gaussian distribution and passed through an MLP with 8 layers. The MLP's weights were masked at each layer based on the generated lower-triangular adjacency matrix, ensuring that connections aligned with the DAG structure. The weights were also randomly sampled and constrained to be positive to enforce monotonicity, with nonlinearity introduced through the leaky ReLU activation function. Additionally, \citep{reisach2021beware} mentioned that the true topological order may be discovered by sorting the marginal variance of each variable. To prevent this phenomenon, the dataset was standardized. 

\textbf{Evaluation metric} \; 
To assess the performance of causal order discovery, we employed the Count Backward $(CB)$ metric, introduced in \citep{rolland2022score}. $CB$ counts the number of backward edges in the ground truth graph with respect to the discovered topological order. This metric quantifies the number of errors in edges of the form $i \rightarrow j$ in the true ground truth graph $\mathcal{G}$, where $\pi_i > \pi_j$ in the discovered topological order $\pi$. $CB$ equals zero when the predicted topological order is correct, and it equals the number of edges in the true graph $\mathcal{G}$ when the predicted topological order is the reverse of the true order.

See Appendix \ref{appendix:algorithm_detail} for a detailed explanation of the architectural design and hyperparameters of Algorithm \ref{alg:proposed} for causal order discovery.

\subsection{Sequential or Permutation based Causal order discovery?}
\textbf{Experimental design} \;
As a baseline, we implemented a procedure for causal order discovery in monotonic SCMs based on learning the permutation matrix for variable ordering and enforcing sparsity of the Jacobian through $l_1$ regularization on the Jacobian of the TMI transformation, following the approach introduced in \citep{xi2023triangular}. For learning the permutation matrix, we used the Gumbel-Sinkhorn network with a temperature parameter $t$ \citep{mena2018learning}. See Appendix \ref{appendix:permutation_detail} for a detailed explanation of the permutation based algorithm. 
As discussed in \citep{xi2023triangular}, satisfying both sparsity and learning the permutation is a complex optimization problem. This challenge is evident in setting the temperature parameter ($t$) and the penalty coefficient of the regularization term ($\lambda$), as it leads to many local minima. In our results, we compare the proposed method with the baseline using different values of $t$ and $\lambda$ to demonstrate the superiority of our approach in terms of optimization efficiency, avoiding many local minima. 

\begin{figure}[h]
    \centering
    \begin{minipage}{0.5\textwidth}
        \centering
        \includegraphics[width=1\textwidth]{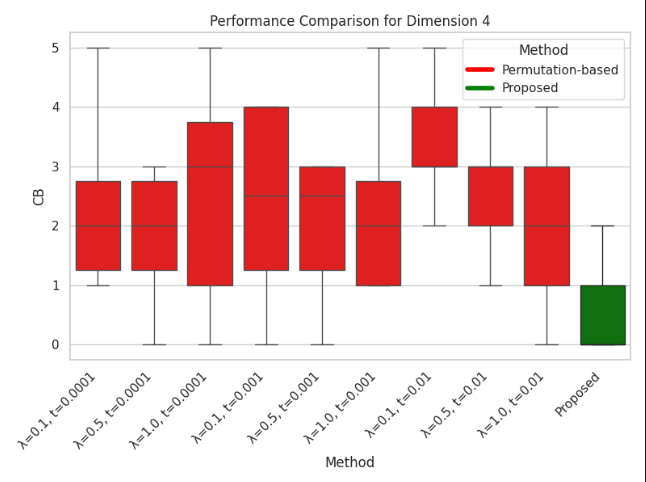} % First image
    \end{minipage}\hfill
    \begin{minipage}{0.5\textwidth}
        \centering
        \includegraphics[width=1\textwidth]{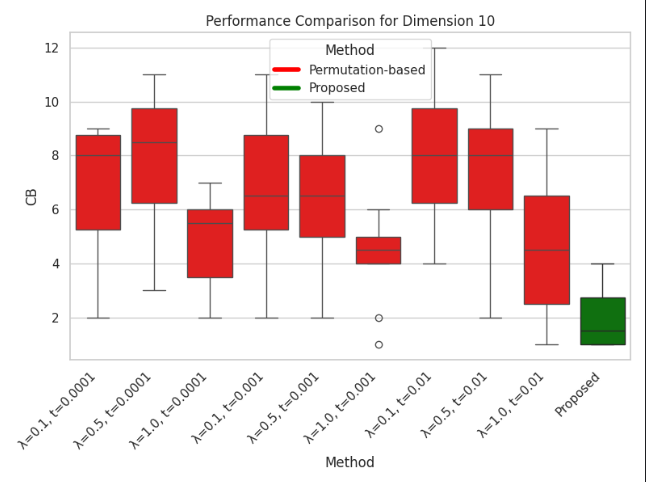} % Second image
    \end{minipage}
    \caption{Performance comparison of sequential order discovery vs permutation-based order discovery}\label{results:task1}
\end{figure}

\textbf{Results}\;
The results for datasets with dimensions 4 and 10 are presented in Figure \ref{results:task1}. Each column in each plot corresponds to a box plot of the given metric over runs of the algorithm for 10 random datasets. The results show that the proposed algorithm consistently outperforms the permutation-based method in terms of Count Backward ($CB$). For both dimensions, different combinations of temperature $t$ and regularization $\lambda$ yield varying results, highlighting the presence of multiple local minima in this approach. In contrast, the proposed method demonstrates superior performance without the need to fine-tune hyperparameters related to ordering and sparsity. 

\subsection{Ablation Analysis}
As discussed in \citep{javaloy2024causal}, we conduct an ablation analysis on the number of flows for both our sequential method and the permutation-based method to balance model complexity and causal consistency. This analysis helps us understand how increasing the number of normalizing flow layers affects causal inconsistency, the likelihood of encountering shortcuts, and the risk of local minima.

\begin{figure}[h]
    \centering
    \begin{minipage}{0.5\textwidth}
        \centering
        \includegraphics[width=1\textwidth]{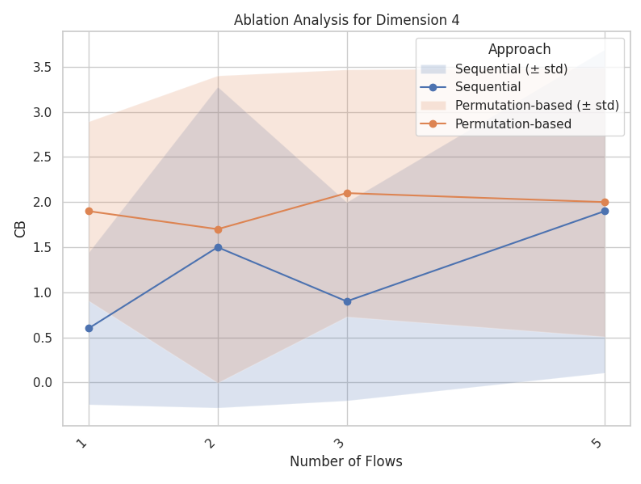} % First image
    \end{minipage}\hfill
    \begin{minipage}{0.5\textwidth}
        \centering
        \includegraphics[width=1\textwidth]{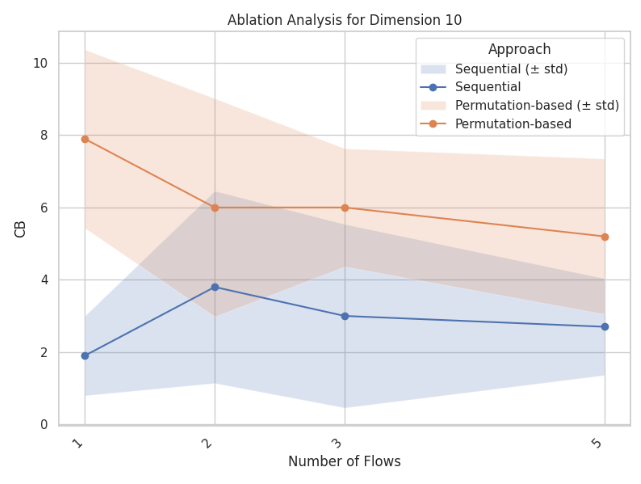} % Second image
    \end{minipage}
    \caption{Ablation analysis on the number of flow layers and hidden layers of MLP.}\label{results:task2}
\end{figure}

Both methods were executed on 10 random synthetic datasets, and the mean and standard deviation of the results from these 10 runs for the \textit{CB} metric are shown in Figure \ref{results:task2}. In the permutation-based approach, the values of $t$ and $\lambda$ are set to 0.0001 and 0.5, respectively. The results indicate that for dimension 10, increasing the number of flow layers results in a higher \textit{CB} for the proposed method, but a lower \textit{CB}  for the permutation-based method. The results for the proposed method are intuitive, as adding more layers of normalizing flow introduces causal inconsistency. However, for the permutation-based method, the performance improves, likely due to the increased complexity of the flow helping to overcome the challenges of the optimization process, as discussed in the previous section.

A similar phenomenon occurred for dimension 4, where increasing the number of flow layers from 1 to 2 improved performance, but the reverse effect was observed with 3 flow layers. This suggests a trade-off between model complexity and causal consistency.

In addition to the above results, detailed evaluations of the proposed method against other causal order discovery approaches, such as SCORE \citep{rolland2022score} and Direct-Lingam \citep{shimizu2011directlingam}, on synthetic and real world datasets are available in Appendix \ref{appendix:results}.

\section{Discussion} 
\textbf{Limitations}
While our method addresses several challenges in causal order discovery for Monotonic SCMs, it assumes that the underlying causal relationships adhere strictly to the monotonicity condition. This assumption may not hold in more complex or real-world scenarios where the relationships between variables are non-monotonic or involve interactions that violate the monotonicity condition. Additionally, our sequential root-identification process, although efficient, may be susceptible to error propagation, especially in high-dimensional settings or when the data is noisy, which could lead to incorrect causal orderings. Additionally, the min-max criterion may not be suitable for high-dimensional settings, as it can overlook variables whose Jacobian values are close to zero. This issue likely arises from the difficulty of accurately estimating the Jacobian using a single-layer normalizing flow. Higher-dimensional cases require more complex functions, but these often struggle with numerous local minima when estimating the Jacobian, leading to large Jacobian values even for the root variable.

\textbf{Future Works}
As the number of variables increases, the computational complexity of iteratively identifying the root variable may become a limiting factor. Future research could focus on developing scalable algorithms to improve efficiency in high-dimensional settings. Another key direction for future work involves increasing the number of normalizing flow layers without compromising causal consistency or getting trapped in local minima. Using more complex normalizing flows can lead to better Jacobian estimation for accurately detecting the root of the graph. Additionally, alternative methods for estimating triangular maps could help overcome the challenges of computing the Jacobian for root variable identification.

Moreover, exploring the benefits of the proposed causal order discovery approach for the overall problem of causal graph discovery would be valuable. One could incorporate different pruning algorithms, such as CAM \citep{buhlmann2014cam}, or leverage identifiability results of monotonic SCMs to extract the adjacency matrix of causal relationships, given the causal graph order, by utilizing a single $d$-dimensional normalizing flow \citep{javaloy2024causal}.

Several studies address the challenges associated with optimization-based algorithms for causal discovery, which can be applied to causal discovery based on monotonic SCMs. For instance, the analysis of sparsity penalties and thresholding discussed in \citep{ng2024structure} provides insights into overcoming the limitations of permutation-based methods. Additionally, ideas such as topological swaps introduced in \citep{deng2023optimizing} can help tackle the difficulties faced in order-based causal discovery. Furthermore, the notion of global optimality proposed in \citep{deng2023global} offers strategies for avoiding local minima in causal discovery methods based on monotonic SCMs.

\section{Conclusion}
In this paper, we introduced a novel sequential procedure for causal order discovery within Monotonic Structural Causal Models (SCMs). Unlike existing methods that rely on complex optimization techniques and sparsity assumptions, our approach iteratively identifies the root variable by leveraging the zeroness of the Jacobian of Triangular Monotonic Increasing (TMI) maps. This method not only simplifies the causal discovery process but also ensures the identification of a unique SCM without requiring multiple independence tests to break the Markov equivalence class. We conducted several experiments comparing our method to those that maximize Jacobian sparsity, demonstrating its efficiency in correctly identifying the causal order.

\bibliographystyle{unsrt}  
\bibliography{Styles/refs}

\medskip

{
\small

\newpage

\section*{Appendix}
\addcontentsline{toc}{section}{Appendix}

%%%%%%%%%%%%%%%%%%%%%%%%%%%%%%%%%%%%%%%%%%%%%%%%%%%%%%%%%%%%

\appendix

\section{Causal order}\label{appendix:causal_order}
For a DAG $\mathcal{G}$, a permutation $\pi : \{1, ..., d\} \rightarrow \{1, ..., d\}$ is a causal topological order if:
\begin{equation}
        \pi (i) < \pi (j) \;\;\; \text{if} \;\;\; j \in DE_i^G 
\end{equation}
where $DE_i^G$ denotes the set of descendents of variable $x_i$ in graph $\mathcal{G}$ \citep{Scholkopf2017-da}. Our proposed algorithm discovers the causal relationships between variables by finding the causal order of variables. We highlight that the causal ordering of a given graph is not necessarily unique \citep{peters2014causal}.

\section{Proof of Theorem \ref{th:root}}\label{appendix:th}

\begin{theorem}
Suppose $X \in \mathcal{R}^{n\times d}$ is generated based on a monotonic SCM according to Equation \ref{eq:mscm}. Then, for the multi-cause model \ref{eq:multi-cause}, $\nabla_{X\backslash\{x_i\}}T_i(x_i, X\backslash\{x_i\})=\mathbf{\vec{0}}$, if and only if $x_i$ is a root in $\mathcal{G}$.
\end{theorem}

\textbf{Proof}; Based on Theorem 1 in \citep{javaloy2024causal}, if the causal order is known, the exogenous variables of the flow differ from the true ones by an invertible component-wise transformation of the variables $u$. Thus, $\nabla_xT(x) = I - A$, where $A$ is a lower triangular matrix. Consequently, the first row corresponds to the root variable, and all values $\nabla T(X)_{1,2:d}$ are zero. However, since the order is unknown, we train each $T_i$ map using a one-dimensional conditional normalizing flow. Instead of conditioning on the preceding variables $x_{1:i}$, we condition on all $X\backslash\{x_i\}$. Because $A$ is lower triangular, $\nabla_{X\backslash\{x_i\}}T_i(x_i, X\backslash\{x_i\})=\mathbf{0}$ for the root variable.

\textbf{Motivating Example}
Suppose we have a monotonic SCM based on the graph $x_1 \rightarrow x_2 \rightarrow x_3$, where $x_1 = f_1(u_1)$, $x_2 = f_2(x_1, u_2)$, and $x_3 = f_3(x_2, u_3)$. Then, we have $u_1 = f^{-1}_1(x_1)$, $u_2 = f^{-1}_2(x_1, x_2)$, and $u_3 = f^{-1}_3(x_2, x_3)$. Since, $\nabla_xf^{-1}(x) = \nabla_xT(x) = I - A$, it follows that:
\begin{equation} 
\nabla_xT(x) = 
\begin{bmatrix}
1 & 0 & 0\\
* & 1 & 0\\
0 & *  & 1\\
\end{bmatrix}
\end{equation}

Thus, the first row corresponds to the root variable, and $\nabla_{x2, x3}f^{-1}_1(x_1, x_2, x_3) = \begin{bmatrix}
   0 & 0  
\end{bmatrix}$ 
while,  $\nabla_{x1, x3}f^{-1}_2(x_2, x_1, x_3) \neq \mathbf{\vec{0}}$ and $\nabla_{x1, x2}f^{-1}_3(x_3, x_1, x_2) \neq \mathbf{\vec{0}}$.

\section{Causal order discovery algorithm}\label{appendix:algorithm}
The algorithm for causal discovery based on Monotonic SCMs works by iteratively identifying the root variable in a dataset. Initially, the set of nodes is all variables, and the order $\pi$ is empty. For each iteration, the algorithm estimates a multi-cause model for each variable, treating it as the effect and the others as its potential causes, using a one-dimensional conditional normalizing flow. The maximum Jacobian value of each variable is computed, and the variable with the smallest maximum Jacobian value is identified as the root. This variable is then added to the causal order $\pi$, and removed from the set of nodes. This process repeats until only one node remains, which is appended to complete the causal order. Algorithm \ref{alg:proposed} shows the overall procedure of the proposed method.

\begin{algorithm}[h]
\caption{
Causal discovery based on Monotonic SCM. 
} \label{alg:proposed}
\begin{algorithmic}[1]
\State \textbf{input}. Data Matrix $X \in \mathcal{R}^{n \times d}$.
\State \textbf{output}. Causal order $\pi$.
\State \textbf{initialize}. $\pi = \emptyset$, nodes = $\{1, ..., d\}$
\Repeat
\For{$i \in nodes$}
 \State estimate multi-cause model \ref{eq:multi-cause}  with causes $X\backslash{x_i}$, and effect $x_i$ using one-dimensional \\ \;\; \; \; \; \; \;conditional normalizing flow. 
 \State $\text{maxJac}(i) = \max \left| \nabla_{X\backslash\{x_i\}} T_i(x_i, X\backslash\{x_i\}) \right| $
\EndFor
 \State $l = \arg \min_{i}\text{maxJac}(i)$
 \State \text{append $l$ to $\pi$}
 \State $nodes \leftarrow nodes - \{l\}$
 \Until{$|$nodes$|=1$}
 \State \text{append $\text{nodes}(1)$ to $\pi$}.
\end{algorithmic}
\end{algorithm}

\section{Proposed Algorithm Detail}\label{appendix:algorithm_detail}
\textbf{Architectural Design} \;
For each $1$-dimensional normalizing flow transformation $T_i(x_i)$ , we employed the neural spline flow \citep{durkan2019neural}, using a stack of 1 flows with 1 hidden block and 128 hidden channels. We also conditioned each $T_i(x_i)$ on $X\backslash\{x_i\}$ using an MLP comprising 1 hidden layer, each with a size of 128. 

The configuration details of the hyperparameters used in our proposed method are summarized in Table \ref{tab:hyperparameters}.

\begin{table}[ht]
    \centering
    \caption{Hyperparameter Configuration for the Proposed Method}
    \begin{tabular}{lc}
        \toprule
        Hyperparameter & Value \\
        \midrule
        Learning Rate & 0.001 \\
        Batch Size & 64 \\
        Number of Epochs & 10 \\
        MLP Dropout Rate & 0.1 \\
        MLP Hidden Layers & 1 \\
        MLP Units per Layer & 128 \\
        Activation Function & Leaky-ReLU \\
        Number of flow layers & 1 \\
        Number of blocks in flow & 1 \\
        Number of hidden channel in flow & 128 \\
        Flow base distribution & Standard Gaussian \\ 
        Optimization Algorithm & Adam (weight decay=0, betas=(0.9. 0.99)) \\
        \bottomrule
    \end{tabular}
    \label{tab:hyperparameters}
\end{table}

\section{Permutation based Algorithm detail}\label{appendix:permutation_detail}
We implement a baseline procedure for causal order discovery in Monotonic SCMs, leveraging a permutation matrix to reorder the variables while regularizing the Jacobian matrix to encourage sparsity. The approach draws upon prior work by \citep{xi2023triangular}. The algorithm follows a two-step process: (1) learning a permutation matrix via the Gumbel-Sinkhorn network, and (2) enforcing sparsity on the Jacobian of the Triangular Monotonic Increasing (TMI) transformation via $l_1$-regularization.

Given an input data matrix $X \in \mathcal{R}^{n\times d}$, our objective is to discover the causal order of variables by learning a permutation matrix $P$, which reorders $X$. The Gumbel-Sinkhorn network is used to generate a soft permutation matrix, which is a continuous relaxation of a discrete permutation matrix. This relaxation enables gradient-based optimization during training. The normalizing flow $T(x.P^T)$ is then trained over the permuted variables, aiming to learn a transformation that captures the causal structure of the data.

The loss function consists of two terms: (1) the log-likelihood of the normalizing flow, and (2) an $l_1$-norm regularization on the Jacobian of the TMI transformation. This regularization enforces sparsity, encouraging the Jacobian to reflect the triangular structure characteristic of SCMs. The loss function is formulated as follows:

\begin{equation}
    \mathcal{L}(\theta, \beta; t, \lambda) := \log p(T_{\theta}^{-1}(u.P_{\beta}(t)^T)) + \lambda||\nabla T_{\theta}(x.P_{\beta}(t)^T)||_1 
\end{equation}

where $\theta$ represents the parameters of the normalizing flow, $\beta$ the parameter of Gumbel-Sinkhorn network, $||\nabla T||_1 $ is the $l_1$-norm applied to the Jacobian matrix to enforce sparsity, weighted by the coefficient $\lambda$, and $P_{\beta}(t)$ refers to the permutation matrix learned via the Gumbel-Sinkhorn network with temperature hyperparameter $t$.

Upon completion of training, the soft permutation matrix can still contain small, non-zero elements that prevent it from being a valid permutation. To address this, we apply the Hungarian algorithm \citep{kuhn1955hungarian} to convert the soft matrix into a hard permutation matrix, ensuring a valid reordering of variables. This final permutation matrix provides the estimated causal order.  Algorithm \ref{alg:baseline} shows the overall procedure of the permutation-based method. To ensure a fair comparison, all shared hyperparameters with the proposed method were set identical to those listed in Table \ref{tab:hyperparameters}.

\begin{algorithm}[h]
\caption{Causal Order Discovery in Monotonic SCMs via Permutation Matrix Learning and Jacobian Sparsity} 
\label{alg:baseline}
\begin{algorithmic}[1]
\State \textbf{Input}: Data matrix $X \in \mathbb{R}^{n \times d}$, temperature parameter $t$, sparsity coefficient $\lambda$
\State \textbf{Output}: Permutation matrix $P$ (causal order)
\State Initialize soft permutation matrix $P_{\beta}(t)$ using the Gumbel-Sinkhorn network with temperature $t$
\Repeat
    \State Train the normalizing flow $T_{\theta}(x.P_{\beta}(t)^\top)$
    \State Compute the Jacobian $\nabla T_{\theta}(x)$
    \State Compute the loss function: 
    \[
    \mathcal{L}(\theta, \beta; t, \lambda) := \log p(T_{\theta}^{-1}(u.P_{\beta}(t)^\top)) + \lambda ||\nabla T_{\theta}(x.P_{\beta}(t)^\top)||_1 
    \]
    \State Update the normalizing flow parameters $\theta$ and the permutation matrix parameters $\beta$ using backpropagation
\Until{convergence of $P_{\beta}(t)$ and $T_{\theta}$}
\State Apply the Hungarian algorithm to convert the soft permutation matrix $P_{\beta}(t)$ into a hard permutation matrix
\State \textbf{return} Hard permutation matrix $P$ (causal order)
\end{algorithmic}
\end{algorithm}

\section{Comparison of Causal Order Discovery}\label{appendix:results}
In this section, we also benchmarked our method against other causal order discovery approaches based on different types of SCMs, such as SCORE \citep{rolland2022score} and RESIT \citep{peters2014causal}, which assume nonlinear additive noise, and Direct-Lingam \citep{shimizu2011directlingam}, which assumes linear SCM. The comparisons were conducted using synthetic datasets containing 1000 samples across varying dimensions, including 2, 3, 4, 5, and 10. For each dimensional setting, we generated ten random datasets to ensure robustness in the evaluation. The experimental results for synthetic datasets are presented in Table \ref{tab:synthetic_results}.

For the assessment of the proposed method on real-world data, we employed dataset of protein-protein interaction networks in the human body (SACHS) \citep{Sachs2005-cu}. This dataset includes ground truth graphs for assessment.  The SACHS dataset comprises measurements of proteins in the human immune system, consisting of $d=11$ nodes and composed of $n=853$ data samples. This dataset's graph contains 17 edges. Additionally, the order of variables has been randomly shuffled to prevent the methods from depending on the variable order. The experimental results for SACHS dataset are presented in Table \ref{tab:sachs_results}.

Our method achieves comparable performance to other causal order discovery approaches across various synthetic datasets, as shown in Table \ref{tab:synthetic_results}, where none of the methods demonstrate a statistically significant advantage. However, for the real-world SACHS dataset, our approach outperforms the alternatives, as seen in Table \ref{tab:sachs_results}. One potential explanation for the challenges observed in synthetic datasets is related to difficulties in accurately estimating the Jacobian. Specifically, in cases where the root variable’s Jacobian value is expected to be zero, errors in this estimation may arise, leading to high Jacobian values, even for the root variable. This issue could stem from limitations in using a single-layer normalizing flow, which may struggle to capture the underlying complexity of the data. Moving forward, increasing the complexity of the normalizing flow, or exploring alternative approaches to estimating triangular maps, could improve Jacobian estimation and lead to more accurate root detection.

\begin{table}[ht]
    \centering
    \caption{Results of different methods on various Synthetic datasets (CB metric)}
    \begin{tabular}{lccccc}
        \toprule
        Method & d=2 & d=3 & d=4 & d=5 & d=10 \\
        \midrule
        Direct-Lingam \citep{shimizu2011directlingam}  & 0.5 $\pm$ 0.53 & 1.0 $\pm$ 0.82 & 1.2 $\pm$ 0.92 & 1.2 $\pm$ 0.79 & 3.0 $\pm$ 2.0 \\
        RESIT \citep{peters2014causal} & 0.2 $\pm$ 0.42 & 0.5 $\pm$ 0.7 & 0.8 $\pm$ 0.79 & 1.1 $\pm$ 1.1 & 2.7 $\pm$ 2.26 \\
        SCORE \citep{rolland2022score}  & 0.3 $\pm$ 0.48 & 0.8 $\pm$ 0.92 & 1.3 $\pm$ 1.49 & 1.6 $\pm$ 1.17 & 3.8 $\pm$ 1.55 \\
        Proposed & 0.2 $\pm$ 0.42 & 1.7 $\pm$ 1.06 & 1.1 $\pm$ 0.74 & 2.0 $\pm$ 1.63 & 3.8 $\pm$ 1.75 \\
        \bottomrule
    \end{tabular}
    \label{tab:synthetic_results}
\end{table}

\begin{table}[ht]
    \centering
    \caption{Results on Real-world dataset SACHS (CB metric)}
    \begin{tabular}{lc}
        \toprule
        Method & SACHS \citep{Sachs2005-cu} \\
        \midrule
        Direct-Lingam \citep{shimizu2011directlingam}  & 8.0 \\
        RESIT \citep{peters2014causal} & 8.0 \\ 
        SCORE \citep{rolland2022score}  & 13.0 \\
        Proposed & 7.0 \\
        \bottomrule
    \end{tabular}
    \label{tab:sachs_results}
\end{table}
\end{document}